%% file: main.tex
\documentclass[times,twocolumn,final,authoryear]{elsarticle}

\usepackage{framed,multirow}

\usepackage{amssymb}
\usepackage{latexsym}

\usepackage{url}
\usepackage{xcolor}
\definecolor{newcolor}{rgb}{.8,.349,.1}

\usepackage[inline]{enumitem}
\usepackage{graphicx}
\usepackage{caption}
\usepackage{subcaption}
\usepackage{arydshln}
\usepackage{todonotes}
\usepackage{pbox}

\usepackage[normalem]{ulem}

\usepackage{multirow}

\input{utils}

\journal{Pattern Recognition Letters}

\begin{document}
\setlength{\abovedisplayskip}{2pt}
\setlength{\belowdisplayskip}{2pt}
\thispagestyle{empty}

\begin{frontmatter}

\title{Graph-Based Neural Network Models with Multiple Self-Supervised Auxiliary Tasks}

\author{Franco Manessi}

\author{Alessandro Rozza}
% \cortext[cor1]{Corresponding author: 
%   Tel.: +39-348-870-2134;  
%   }
% \ead{alessandro.rozza@lastminute.com}

% \address[1]{
%   Strategic Analytics, 
%   lastminute.com group,
%   Chiasso, Switzerland, 
%   \texttt{\{first\_name\}.\{last\_name\}@lastminute.com}}

% % \received{1 May 2013}
% % \finalform{10 May 2013}
% % \accepted{13 May 2013}
% % \availableonline{15 May 2013}
% \communicated{F. Manessi}

\begin{abstract}
  Self-supervised learning is currently gaining a lot of attention, as it allows neural networks to learn robust representations from large
  quantities of unlabeled data. Additionally, multi-task learning can further improve representation learning by training networks simultaneously
  on related tasks, leading to significant performance improvements. 
  In this paper, we propose three novel self-supervised auxiliary tasks to train graph-based neural network models in a
  multi-task fashion.
  Since Graph Convolutional Networks are among the
  most promising approaches for capturing relationships among structured data points, we use them as a building block to achieve competitive results
  on standard semi-supervised graph classification tasks.
\end{abstract}

% \begin{keyword}
% \MSC 41A05\sep 41A10\sep 65D05\sep 65D17
% \KWD Keyword1\sep Keyword2\sep Keyword3

% \end{keyword}

\end{frontmatter}

%\linenumbers
\input{introduction}
\input{related}
\input{methods}

\input{results}

\input{conclusion}

\subsubsection*{Acknowledgments}
The authors would like to thank Adam Elwood for his helpful and constructive comments that contributed to improve the
work.

\bibliographystyle{model2-names}
\bibliography{bibliography}

\end{document}

%% file: utils.tex
\usepackage{xfrac}
\usepackage{bm}
\usepackage{xspace}
\usepackage{mathrsfs}
\usepackage{amsmath,amsfonts,bm}
\usepackage{mathtools}

\newcommand\Matrix[1]{\bm{#1}}
\newcommand\MatrixElements[1]{{#1}}
\newcommand\MatrixSpace[2]{\mathbb{R}^{#1\times#2}}
\newcommand\Graph[1]{\bm{\mathcal{#1}}}
\newcommand\ie{i.e.\xspace}
\newcommand\eg{e.g.\xspace}
\newcommand\Reals{\mathbb{R}}
\newcommand\Id{\Matrix{I}}
\DeclareMathOperator{\relu}{ReLU}
\DeclareMathOperator{\softmax}{softmax}
\newcommand\vect[1]{{\boldsymbol{#1}}}

\newcommand\Set[1]{{\bm{\mathcal{#1}}}}
\newcommand\muacro[1]{\texttt{#1}} 
\newcommand\card[1]{{|#1|}}
\newcommand\Minus{\textrm{-}}
\newcommand\GCN{\muacro{GCN}\xspace}
\newcommand\GC{\muacro{GC}\xspace}

\def\eqref#1{equation~(\ref{#1})}
\def\Eqref#1{Equation~(\ref{#1})}

%% file: introduction.tex
\section{Introduction}
In the last decade, neural networks approaches that can deal with with structured data have been gaining a lot of
traction \citep{Scarselli09, Bruna13, defferrard2016convolutional, kipf2016semi,Manessi20}. Due to the prevalence of
data structured in the form of graphs, the capability to explicitly exploit structural relationships among  data points
is particularly useful in improving the performance for a variety of tasks. Graph Convolutional Networks ({\GCN}s,
\citet{kipf2016semi}) stand out as a particularly successful iteration of such networks, especially for semi-supervised
problems. {\GCN}s act to encode graph structures, while being trained on a supervised target loss for all the nodes with
labels. This technique is able to share the gradient information from the supervised loss through the graph adjacency
matrix and to learn representations exploiting both labeled and unlabeled nodes. Although {\GCN}s can stack multiple
graph convolutional layers in order to capture high-order relations, these architectures suffer from ``over-smoothing''
when the number of layers increases \citep{li2018deeper}, thus making difficult to choose an appropriate number of
layers.

If we have a dataset with enough labels, supervised learning can usually achieve good results. Unfortunately, to label a
large amount of data is an expensive task. In general, the amount of unlabelled data is substantially more than the data
that has been human curated and labelled. It is therefore valuable to find ways to make use of this unlabelled data.
% However, unsupervised learning usually works worse than supervised learning. 
A potential solution to this problem comes if we can get labels from unlabelled data and train unsupervised dataset in a
supervised manner. Self-supervision achieves this by automatically generating additional labelled signals from the
available unlabelled data, using them to learn representations. A possible approach in deep learning involves taking a
complex signal, hiding part of it from the network, and then asking the network to fill in the missing information
\citep{Doersch17}.

% Another way to overcome the lack of labeled data can be found whenever there are multiple related tasks each of which
% has limited training samples. Indeed, it is found that learning different tasks jointly can lead to improve the
% performance compared with learning them individually, if all of them or at least a subset of them are assumed to be
% related to each other \citep{caruana1997multitask}.

Additionally, it is found that joint learning of different tasks can improve performance over learning them
individually, given that at least a subset of these tasks are related to each other \citep{caruana1997multitask}. This
observation is at the core of multi-task learning. Precisely, given $T$ tasks $\{\bm{\mathcal{T}}_{i}\}^T_{i=1}$ where a
subset of them are related, multi-task learning aims to help improve the learning of a model for
$\{\bm{\mathcal{T}}_{i}\}^T_{i=1}$ by using the knowledge contained in all or some of the $T$ tasks (\cite{Zhang17}).

In this paper we train neural network-based graph architectures by means of self-supervised auxiliary tasks in a
 multi-task framework, similarly to \citet{you2020does}. Considering the promising results of the \GCN, we decided to
 experiment this framework in semi-supervised classification problems on graphs, employing \GCN as a base building
 block. The main contribution of this paper consists of three novel auxiliary tasks for graph-based neural networks:
\begin{description}[itemsep=1pt,parsep=1pt,topsep=0pt, partopsep=0pt]

    \item[autoencoding:] with which we aim at extracting node representations robust enough to allow both
        semi-supervised classification as well as vertex features reconstruction;

    \item[corrupted features reconstruction:] with which we try to extract node representations that allows to
        reconstruct some of the vertex input features, starting from an embedding built from a corrupted version of
        them. This auxiliary task can be seen as the graph equivalent of reconstructing one of the color channels of a
        RGB image using the other channels in computer vision self-supervised learning;

    \item[corrupted embeddings reconstruction:] with which we try to extract node representations robust to embedding
        corruption. This is similar to the aforementioned auxiliary task, with the difference that the reconstruction is
        performed on the node embeddings instead of the vertex features.

\end{description}

These three tasks are intrinsically self-supervised, since the labels are directly extracted from the input graph and
its vertex features. These novel auxiliary tasks allow to achieve competitive results on standard datasets and to reduce
the aforementioned ``over-smoothing'' limitation of deep {\GCN}s.

The paper is organized as follows: in Section 2 the related works are summarized; in Section 3 we introduce the three
auxiliary tasks; in Section 4 a detailed comparison against \GCN on a standard public datasets is presented; Section 5
reports conclusions and future works.

%% file: related.tex
\section{Related works}

In recent years, graph representation learning have gained a lot of attention. These techniques can be divided in three
main categories: 
\begin{enumerate*}[label=(\roman*)]
    \item random walk-based;
    \item factorization-based;
    \item neural network-based.
\end{enumerate*}
In the first group, \emph{node2vec} \citep{Grover16} and \emph{Deepwalk} \citep{Perozzi14} are worth mentioning. The
former is an efficient and scalable algorithm for feature learning that optimizes a novel network-aware, neighborhood
preserving objective function, using stochastic gradient descent. The latter uses truncated random walks to efficiently
learn representations for vertices in graphs. These latent representations, which encode graph relations in a vector
space, can be easily exploited by standard statistical models to produce state-of-the-art results.

Among the factorization based methods, \citet{Xu13} presents a semi-supervised factor graph model that can exploit the
relationships among the nodes. In this approach, each vertex is modeled as a variable node and the various relationships
are modeled as factor nodes.

In the last group, we find all the works that have revisited the problem of generalizing neural networks to work on
structured graphs, some of them achieving promising results in domains that have been previously dominated by other
techniques. \citet{Gori2005} and \citet{Scarselli09} formalize a novel neural network model, the Graph Neural Network.
This model maps a graph and its nodes into a $D$-dimensional Euclidean space in order to learn a final
classification/regression model. \citet{Bruna13} approach the graph structured data by proposing two generalizations of
Convolutional Neural Networks (\muacro{CNN}s): one based on a hierarchical clustering of the domain and another based on
the spectrum of the graph (computed using the Laplacian matrix). \citet{defferrard2016convolutional} extend the spectral
graph theory approach of the previous work by providing efficient numerical schemes to design fast localized
convolutional filters on graphs, achieving the same computational complexity of classical \muacro{CNN}s.
\citet{kipf2016semi} build on this idea by introducing {\GCN}s. They exploit a localized first-order approximation of
the spectral graph convolutions framework \citep{Hammond11}. Recently, \citet{gat} have applied attention mechanism to
graph neural networks to improve model performance.

However, the majority of the methods belonging the three aforementioned categories require a large amount of labelled
data, which can limit their applicability. On the other hand, unsupervised algorithms, such as \citet{NIPS2017_6703,
pmlr-v97-grover19a, velickovic2019deep} do not require any external labels, but their performances usually suffer when
compared to supervised techniques.

Self-supervised learning can be considered a branch of unsupervised learning, where virtually unlimited supervised
signals are generated from the available data and used to learn representations. This learning framework finds many
applications, ranging from language modeling \citep{wu2019self, 2013arXiv1301.3781M, radford2018improving}, to robotics
\citep{jang2018grasp2vec}, and computer vision \citep{zhang2016colorful, ledig2017photo, pathak2016context,
zhang2017split, noroozi2016unsupervised, doersch2015unsupervised}. Applied to graph representation learning,
\citet{sun2020multi} proposed a multi-stage self-supervised framework, called \muacro{M3S}, showing some empirical
success.

Multi-task learning approaches can be divided in five many categories:
\begin{enumerate*}[label=(\roman*)]
    \item feature learning;
    \item low-rank approaches; 
    \item task clustering;
    \item task relation learning;
    \item decomposition \end{enumerate*} \citep{zhang2017survey}. In the feature learning approach, it is assumed that
different tasks share a common feature representation based on the original features. In the Multi-Task Feature Learning
method, task specific hidden representations within a shallow network are obtained by learning the feature covariance
for all the tasks, in turn allowing to decouple the learning of the different tasks \citep{argyriou2007multi,
argyriou2008convex}. A common approach applied in the deep learning setting is to have the different tasks share the
first several hidden network layers, including task-specific parameters only in the subsequent layers
\citep{zhang2014facial, mrkvsic2015multi, li2014heterogeneous}. A more complex approach in deep learning is the
cross-stitch network, proposed by \citet{misra2016cross}, in which each task has its own independent hidden layers that
operate on learned linear combinations of the activation maps of the previous layers.

The low-rank approaches assume that the model parameters of different tasks share a low-rank subspace
\citep{ando2005framework}. \citet{pong2010trace} propose to regularize the model parameters by means of the trace norm
regularizer, in order to exploit the property of the trace norm to induce low rank matrices. The same idea has been
applied in deep learning by \citet{yang2016trace}.

Another approach is to assume that different tasks form several clusters, each of which consists of similar tasks. This
can be thought of as clustering algorithms on the task level, while the conventional clustering algorithms operate on
the data level. \citet{thrun1996discovering} introduced the first implementation of this idea for binary classification
tasks that are defined over the same input space. \citet{bakker2003task} followed the idea to recast the neural networks
used in the feature learning approach in a Bayesian settings, where the weights of the task specific final layers are
assumed to have a Gaussian mixture as a prior. Similarly, \citet{xue2007multi} build on the previous idea by changing
the prior to a Dirichlet process.

In the task relation based approaches, the task relatedness (\eg task correlation or task covariance) is used to drive
the joint training of multiple tasks. In the early works, these relations are assumed to be known in advance. They are
used to design regularizers to guide the learning of multiple tasks, so that the more similar two tasks are, the closer
the corresponding model parameters are expected to be \citep{evgeniou2005learning,kato2008multi}. However, task
relations are often not available and need to be automatically estimated from data. \citet{bonilla2008multi} go into
this direction by exploiting Gaussian processes and defining a multivariate normal prior on the functional output of all
the task outputs, whose covariance is trained from data and represents the relation between the tasks.

In the decomposition approach, it is assumed that the matrix whose row vectors are the weights of each of the tasks can
be decomposed as a linear combination of two or more sub-matrices, where each sub-matrix is suitably regularized
\citep{jalali2010dirty, chen2012learning, zhong2012convex}.

The first attempt to combine self-supervision and multi-task learning on top of {\GCN}s can be found in
\citep{you2020does}. The authors compare the direct usage of self-supervision against self-supervision by means of
multi-task learning, showing that the latter approach achieves better results. In their paper, they introduce three
self-supervised auxiliary tasks, \ie node clustering, graph partitioning, and graph completion. It is important to 
notice that these auxiliary tasks are very different with respect to the ones introduced in this paper.

%% file: methods.tex
\section{Methods}\label{sec:methods}
In this section, we introduce the formalization of a multi-task self-supervised \GCN for semi-supervised classification.
We will first give some preliminary definitions, including of a Graph Convolutional (\GC) layer and  multi-task target
loss. We then proceed by showing the auxiliary tasks that can be learned jointly with the semi-supervised classification
loss. Finally, we introduce the overall architecture we used in our experiments.

\subsection{Preliminaries}\label{sec:preliminaries} 
Let $\MatrixElements{Y}_{i,j}$ be the $i$-th row, $j$-th column element of the matrix $\Matrix{Y}$. $\Matrix{I}_d$ is
the identity matrix in $\Reals^d$; $\softmax$ and $\relu$ are the \emph{soft-maximum} and the \emph{rectified linear
unit} activation functions \citep{goodfellow2016deep}. Note that all the activation functions act element-wise when
applied to a matrix.

An \emph{undirected} graph $\Graph{G} = (\Set{V}, \Set{E})$ is defined by its set of the nodes (or vertices), $\Set{V}$,
and set of the edges, $\Set{E}$. For each vertex $v_i \in \Set{V}$ let $\vect{v}_i \in \Reals^d$ be the corresponding
feature vector. Moreover, let $\Matrix{A}$ be the adjacency matrix of the graph $\Graph{G}$; namely, $\Matrix{A} \in
\MatrixSpace{\card{\Set{V}}}{\card{\Set{V}}}$ where $\MatrixElements{A}_{i,j} = \MatrixElements{A}_{j,i} = w_{ij}$ if
and only if there is an edge between the $i$-th and $j$-th vertices and the edge has weight $w_{ij}$. In the case of an
unweighted graph, $w_{ij} = 1$. The symbol $\Matrix{X}$ will denote instead the vertex-features matrix $\Matrix{X} \in
\MatrixSpace{\card{\Set{V}}}{d}$, \ie the matrix whose row vectors are the $\vect{x}_i$.

The mathematics of the \muacro{GC} layer \citep{kipf2016semi} is here briefly recalled, since it is a basic building
block of the following network architectures. Given a graph with adjacency matrix
$\Matrix{A}\in\MatrixSpace{\card{\Set{V}}}{\card{\Set{V}}}$ and vertex-feature matrix
$\Matrix{X}\in\MatrixSpace{\card{\Set{V}}}{d}$, the \muacro{GC} layer with $M$ output nodes (also called channels) and
$\Matrix{B}\in\MatrixSpace{d}{M}$ weight matrix is defined as the function $\operatorname{GC}_{M}$ from
$\MatrixSpace{\card{\Set{V}}}{d}$ to $\MatrixSpace{\card{\Set{V}}}{M}$ such as $\operatorname{GC}_{M}(\Matrix{X})
\coloneqq \hat{\Matrix{A}}\Matrix{X}\Matrix{B}$, where $\hat{\Matrix{A}}$ is the re-normalized adjacency matrix, \ie
$\hat{\Matrix{A}} \coloneqq \tilde{\Matrix{D}}^{\,\Minus\sfrac{1}{2}} \tilde{\Matrix{A}}
\tilde{\Matrix{D}}^{\,\Minus\sfrac{1}{2}}$ with $\tilde{\Matrix{A}} \coloneqq \Matrix{A} + \Id_{\card{\Set{V}}}$ and
$\tilde{\MatrixElements{D}}_{kk}\coloneqq \sum_l \tilde{\MatrixElements{A}}_{kl}$. Note that the \muacro{GC} layer can
be seen as localized first-order approximation of spectral graph convolution \citep{defferrard2016convolutional}, with
the additional \emph{renormalization trick} in order to improve numerical stability \citep{kipf2016semi}.

Consider now a multi-task problem, made of $T$ tasks, indexed by $t=1,\ldots,T$. All the tasks share the input space
$\Set{X}$ and have the task-specific output spaces $\Set{Y}_t$. We suppose that each task $t$ is associated to a
parametric hypothesis class $f_t$ (\eg a neural network architecture) such that $f_t(\vect{x}; \vartheta_\text{sh},
\vartheta_t) = \hat y^{t}$, where $\vect{x} \in \Set{X}$, $y^t \in \Set{Y}_t$, $\vartheta_\text{sh}$ is a parameter
vector shared among the hypothesis classes of different tasks, and $\vartheta_t$ is task-specific. The joint training of
each of the $f_t$ is achieved by means of empirical risk minimization:
% \vspace{-1.5em}
\begin{equation}\label{eq:multi_task_loss}
    \operatorname{argmin}_{\vartheta_\text{sh}, \vartheta_1, \ldots, \vartheta_T} 
        \sum_{t=1}^{T} w_t \mathcal{R}_t(\vartheta_\text{sh}, \vartheta_t),
        % \vspace{-1.5em}
\end{equation}
where $w_t \in \Reals^+$ and $\mathcal{R}_t(\vartheta_\text{sh}, \vartheta_t)$ are the task-specific empirical risks.
Precisely, $\mathcal{R}_t(\vartheta_\text{sh}, \vartheta_t) \coloneqq \frac{1}{N} \sum_i \mathcal{L}_t(f_t(\vect{x}_i;
\vartheta_\text{sh}, \vartheta_t), y_i^t)$ with $\mathcal{L}_t$ the task-specific loss function, $\vect{x}_i$ the
feature vectors of the $i$-th training sample, $y_i^t$ the target variable of the $i$-th training sample corresponding
to the $t$-th task, and $N$ the total number of training samples. Roughly speaking, the multi-task objective of
\Eqref{eq:multi_task_loss} is the conic combination with weights $w_t$ of the empirical risk of each task. A basic
justification for taking the weighted combination is due to the fact that it is not possible to define global optimality
in the multi-task setting. Indeed, consider two sets of solutions $(\vartheta_\text{sh}, \vartheta_1, \vartheta_2)$ and
$(\bar\vartheta_\text{sh}, \bar\vartheta_1, \bar\vartheta_2)$ such that $\mathcal{R}_1(\vartheta_\text{sh}, \vartheta_1)
< \mathcal{R}_1(\bar\vartheta_\text{sh}, \bar\vartheta_1)$ and $\mathcal{R}_2(\vartheta_\text{sh}, \vartheta_2) >
\mathcal{R}_2(\bar\vartheta_\text{sh}, \bar\vartheta_2)$, \ie $(\vartheta_\text{sh}, \vartheta_1, \vartheta_2)$ is the
best solution for the first task, while $(\bar\vartheta_\text{sh}, \bar\vartheta_1, \bar\vartheta_2)$ reaches optimality
in the second task. It is not possible to compare these two solutions without a pairwise measure. A way to put
them on the same footing is by mean of \Eqref{eq:multi_task_loss}. 

The weights $w_t$ will be considered as static hyper-parameters of the training procedure in the remaining of the paper.
It is worth mentioning that also other approaches exist in which the weights are dynamically computed or obtained
through an heuristic \citep{chen2018gradnorm, kendall2018multi}.

It is worth noting that the framework we are considering is usually called \emph{hard parameter sharing}, \ie there
are some parameters $\vartheta_\text{sh}$ that are shared among all the tasks. On the other hand, in \emph{soft
parameter sharing}, all parameters are task-specific but they are jointly constrained by means of regularization.

\subsection{The tasks}\label{sec:tasks}
This section is organized as follows: in \ref{sec:main-task} the main task is defined; in \ref{sec:autoencoding},
\ref{sec:cfr}, \ref{sec:cer} the auxiliary tasks are formalized.

\subsubsection{The main task}\label{sec:main-task}
As mentioned before, we will consider the semi-supervised classification of graph nodes as our main task. However, what
follows can easily be extended to other main tasks as well.

Let's consider a $K$-class semi-supervised classification problem; thus the output space of the main task can be written
as $\Set{Y}_\text{main} \coloneqq \{ \vect{y} \in \Reals^K \mid y_k \in \{0, 1\},\ \sum_k y_k = 1 \}$, \ie the space of
one-hot encoded $K$-class vectors. By denoting with $\Set{V}_\text{l} \subseteq \Set{V}$ the subset of the labeled nodes
of the graph $\Graph{G}$, the empirical risk $\mathcal{R}_\text{main}$ of the main task, corresponding to a
cross-entropy loss, can be written as:
\begin{equation*}
    \mathcal{R}_\text{main} \coloneqq \frac{1}{\card{\Set{V}_\text{l}}} \sum_{i\in\Set{V}_\text{l}}
        \sum_{k=1}^{K} y_k \log f_\text{main}(\vect{x}_i; \vartheta_\text{sh}, \vartheta_\text{main}),
\end{equation*}
with $0 \times \log 0 = 0$. We make the assumption that $f_\text{main} \coloneqq g_\text{sh} \circ g_\text{main}$, with $ \partial g_\text{sh} /
\partial \vartheta_\text{main} = \partial g_\text{main} / \partial \vartheta_\text{sh} = 0$, namely, $f_\text{main}$ can
be seen as the function composition of a vertex feature embedding function $g_\text{sh}$ parameterized only by
$\vartheta_\text{sh}$, followed by a task specific classification head $g_\text{main}$ parameterized by
$\vartheta_\text{main}$ only. As we will see later, $g_\text{sh}$ is shared with the auxiliary tasks. Finally,
$g_\text{sh}$, $g_\text{main}$, and all the functions we will discuss further ahead are considered differentiable almost
everywhere.

\subsubsection{Autoencoding}\label{sec:autoencoding}
The objective in the autoencoding task (also called \muacro{AE}) is to reconstruct the graph vertex features from an
encoding thereof. Using the mean squared error reconstruction loss, the corresponding empirical risk
$\mathcal{R}_\text{AE}$ can be written as:
\begin{equation*}
    \mathcal{R}_\text{AE} \coloneqq \frac{1}{\card{\Set{V}_\text{AE}}} \sum_{\substack{
    i\in\Set{V}_\text{AE}\\
    \Set{V}_\text{AE}\subseteq\Set{V}
    }}\!\!
        \lVert \vect{x}_i - f_\text{AE}(\vect{x}_i; \vartheta_\text{sh}, \vartheta_\text{AE}) \rVert_2^2,
\end{equation*}
with $f_\text{AE} \coloneqq g_\text{sh} \circ g_\text{AE}$, 
${\partial g_\text{AE}}/{\partial \vartheta_\text{sh}} = 0$. Namely, the autoencoder is made of the encoder function $g_\text{sh}$ also present in the main task, and a decoder
component specified by $g_\text{AE}$ that depends on the task specific parameters $\vartheta_\text{AE}$ only.

\subsubsection{Corrupted features reconstruction}\label{sec:cfr}
The aim of this task (also called \muacro{FR}) is to reconstruct the graph vertex features from an encoding of a
corrupted version of them. Namely, the goal is to train an autoencoder that it is able to restore vertex features
starting from a vertex-feature matrix $\Matrix{X}$ that has some columns zeroed out, \ie corrupted.

We distinguish two methods:
\begin{enumerate*}[label=(\roman*)]
    \item \emph{partial reconstruction}, where we aim at outputting the restored features only;
    \item \emph{full reconstruction}, where we aim at outputting also the non-corrupted ones.
\end{enumerate*}

Let $\Set{M}$ be the subset $\Set{M} \subset \{1, \ldots, d \}$, and $\Matrix{P}_\Set{M}  \in \MatrixSpace{d}{d}$ the
diagonal matrix such as its $i$-th diagonal elements are $1$ for all $i\notin \Set{M}$, and $0$ otherwise, \ie
$\Matrix{P}_\Set{M}$ is the identity matrix with some elements equal to zero. When applied to a column vector $\vect{v}
\in \Reals^d$, such a matrix has the property of zero-ing out all the vector elements corresponding to the indexes
belonging to $\Set{M}$. %, \ie it is a projection matrix.

Thanks to $\Matrix{P}_\Set{M}$, and considering the mean squared error reconstruction loss, the empirical risk
$\mathcal{R}_\text{FR}^\text{f}$ corresponding to the corrupted \emph{full} features reconstruction can be written as:
\begin{equation}\label{eq:cfr-full-task-empirical-risk}
    \mathcal{R}_\text{FR}^\text{f} \coloneqq \frac{1}{\card{\Set{V}_\text{FR}}} \sum_{\substack{
    i\in\Set{V}_\text{FR}\\
    \Set{V}_\text{FR}\subseteq\Set{V}
    }}\!\!
    \lVert \vect{x}_i - f_\text{FR}^\text{f}(
        \Matrix{P}_\Set{M} \vect{x}_i; \vartheta_\text{sh}, \vartheta_\text{FR}^\text{f}
    ) \rVert_2^2,
\end{equation}
with $f_\text{FR}^\text{f} \coloneqq g_\text{sh} \circ g_\text{FR}^\text{f}$, 
${\partial g_\text{FR}^\text{f}}/{\partial \vartheta_\text{sh}} = 0$.
Namely, $f_\text{FR}^\text{f}$ acts as a \emph{denoising} autoencoder, with the input corrupted by the matrix
$\Matrix{P}_\Set{M}$ (for some arbitrary chosen $\Set{M}$), and as encoder the function $g_\text{sh}$. 
The decoder component is specified by $g_\text{FR}^\text{f}$, that depends on task specific parameters
$\vartheta_\text{FR}^\text{f}$ only.

Now, we will consider the partial features reconstruction. Let $\Matrix{I}_\Set{M} \in \MatrixSpace{\card{\Set{M}}}{d}$
be a rectangular matrix whose $i$-th row is a zero vector, with only a $1$ at the $j$-th position, with $j\in\Set{M}$.
When applied to a column vector $\vect{v} \in \Reals^d$, such a matrix has the property of selecting the vector elements
corresponding to the indexes belonging to $\Set{M}$. Leveraging $\Matrix{I}_\Set{M}$, the empirical risk
$\mathcal{R}_\text{FR}^\text{p}$ corresponding to the corrupted \emph{partial} features reconstruction can be written
similarly as:
\begin{equation}\label{eq:cfr-part-task-empirical-risk}
    \mathcal{R}_\text{FR}^\text{p} \coloneqq \frac{1}{\card{\Set{V}_\text{FR}}} \sum_{\substack{
    i\in\Set{V}_\text{FR}\\
    \Set{V}_\text{FR}\subseteq\Set{V}
    }}\!\!
    \lVert \Matrix{I}_\Set{M} \vect{x}_i - f_\text{FR}^\text{p}(
        \Matrix{P}_\Set{M} \vect{x}_i; \vartheta_\text{sh}, \vartheta_\text{FR}^\text{p}
    ) \rVert_2^2,
\end{equation}
with $f_\text{FR}^\text{p} \coloneqq g_\text{sh} \circ g_\text{FR}^\text{p}$, 
${\partial g_\text{FR}^\text{p}}/{\partial\vartheta_\text{sh}} = 0$.

\subsubsection{Corrupted embeddings reconstruction}\label{sec:cer}
Similarly to the previous task, the aim is to reconstruct ``something'' from a corrupted version of it. In this case
(also called \muacro{ER}), the goal is to reconstruct the embeddings produced by some encoder, in order to make the
embeddings resilient to noise. Also in this case, the corruption is achieved by zero-ing out some entries,
distinguishing two methods:
\begin{enumerate*}[label=(\roman*)]
    \item \emph{partial reconstruction}, where we aim at outputting the restored embeddings only;
    \item \emph{full reconstruction}, where we aim at outputting the restored embeddings as well as the non-corrupted
        ones.
\end{enumerate*}

Considering the full reconstruction case, the mean squared error loss, and $\Set{N}$ as the set containing the corrupted
embedding index, we can write the empirical risk $\mathcal{R}_{ER}^\text{f}$ corresponding to the corrupted \emph{full}
embeddings reconstruction can be written as:
\begin{equation}\label{eq:cer-full-task-empirical-risk}
    \mathcal{R}_\text{ER}^\text{f} \coloneqq \frac{1}{\card{\Set{V}_\text{ER}}} \sum_{\substack{
    i\in\Set{V}_\text{ER}\\
    \Set{V}_\text{ER}\subseteq\Set{V}
    }}\!\!
    \lVert g_\text{sh}(\vect{x}_i) - f_\text{ER}^\text{f}(
        \vect{x}_i; \vartheta_\text{sh}, \vartheta_\text{ER}^\text{f}
    ) \rVert_2^2,
\end{equation}
with $f_\text{ER}^\text{f} \coloneqq 
g_\text{sh} \circ \Matrix{P}_\Set{N} \circ g_\text{ER}^\text{f}$, 
${\partial g_\text{ER}^\text{f}}/{\partial \vartheta_\text{sh}}\ \  = 0$. Namely, we use the function $g_\text{sh}$ also present in the main task as our encoder producing vertex embeddings, we
corrupt them by zero-ing out some of them with $\Matrix{P}_\Set{N}$, and we try to reconstruct them with the decoder
defined by $g_\text{ER}^\text{f}$.

The \emph{partial} reconstruction version can be obtained by leveraging $\Matrix{I}_\Set{N}$ as made in
Section~\ref{sec:cfr}.

\subsection{The final network}\label{sec:final-network}
\begin{figure}[t]
    \centering
    \includegraphics[width=.9\columnwidth]{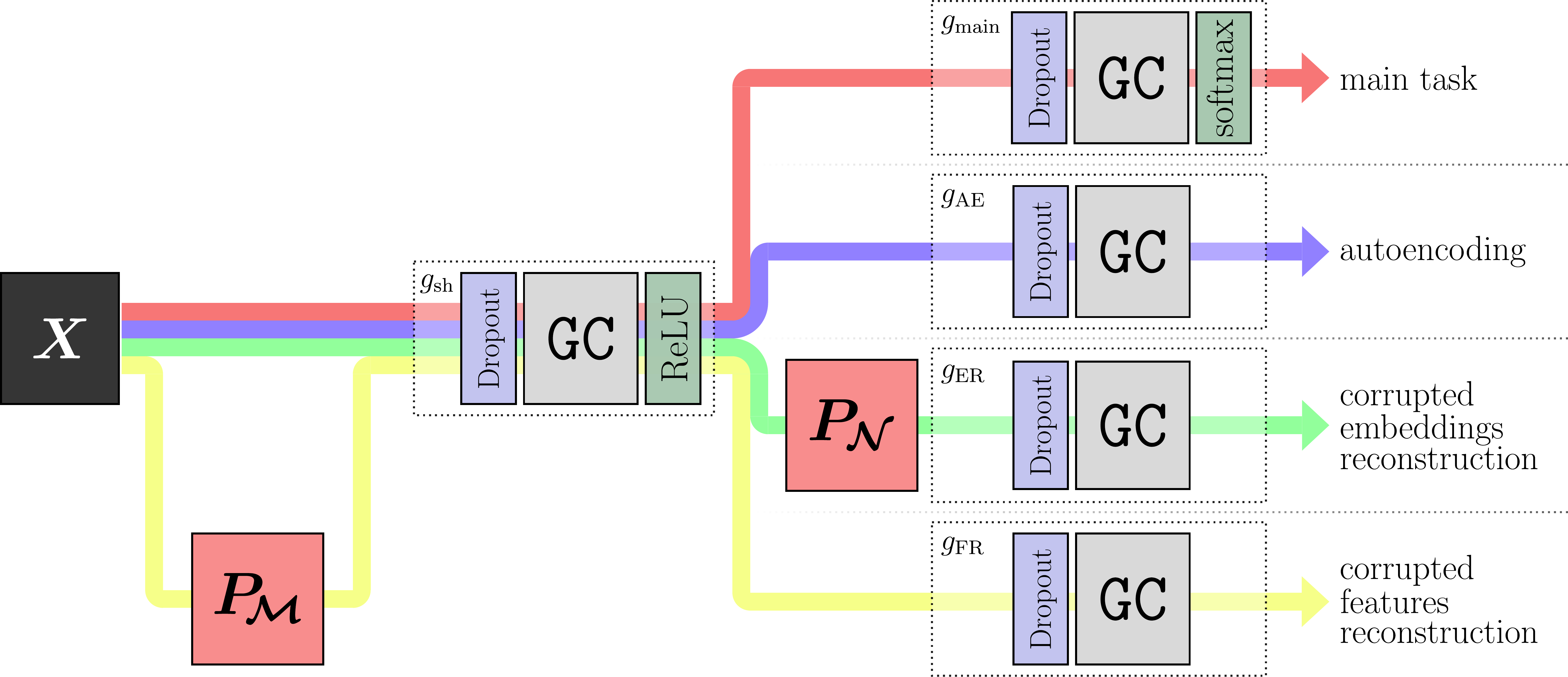}
    \caption{\label{fig:final-network}
        The drawing shows the architecture of the network described in Section~\ref{sec:final-network}, which is the
        one used in our experiments. The network is made of a shared encoder $g_\text{sh}$, followed by four heads,
        one devoted to the main task $g_\text{main}$, and the other three to each of the auxiliary tasks $g_\text{AE}$,
        $g_\text{FR}$, $g_\text{ER}$.
    }
    \vspace{-1em}
\end{figure}

% The formalism introduced in the previous section does not make any assumptions about the mathematical form of the
% various encoder and decoder functions. This means that the functions $g_\text{sh}$, $g_\text{main}$, $g_\text{AE}$,
% $g_\text{FR}$, $g_\text{ER}$ can be modeled by an arbitrary neural network, with any kind of layers, number of units,
% and activation functions. In the rest of the paper, we will restrict our analysis by using as a foundation block the \GC
% layer \citep{kipf2016semi}.

Our overall network, is composed of a shared encoder $g_\text{sh}$, and four output heads $g_\text{main}$,
$g_\text{AE}$, $g_\text{FR}^\text{f}$, $g_\text{ER}^\text{f}$, one per task. Note that we are going to present
explicitly only the \emph{full reconstruction} variant of the network, since the \emph{partial reconstruction} version
can be easily derived. In the rest of the paper, we will restrict our analysis by using as a foundation block the \GC
layer \citep{kipf2016semi}.

The shared encoder $g_\text{sh}$ is the same used in \citet{kipf2016semi}, \ie a dropout layer
\citep{srivastava2014dropout} with 50\% dropout rate followed by  \GC layer made of 16 units and a $\relu$ activation
function: $g_\text{sh} = \operatorname{Dropout}(0.5) \circ \operatorname{GC}_{16} \circ \relu$. The main task
classification head $g_\text{main}$ is made of a dropout layer followed by a \GC layer and a $\softmax$ activation,
where the units of the \GC layer depends on the number of classification classes ($g_\text{main} =
\operatorname{Dropout}(0.5) \circ \operatorname{GC} \circ \softmax$).

All the auxiliary task heads $g_\text{AE}$, $g_\text{FR}^\text{f}$, $g_\text{ER}^\text{f}$ are made of a dropout layer
followed by a \GC layer made of 16 units, a $\relu$ activation function, another dropout layer, and a final \GC layer
with no activation function: $\operatorname{Dropout}(0.5) \circ \operatorname{GC}_{16} \circ \relu \circ
\operatorname{Dropout}(0.5) \circ \operatorname{GC}$. Note that the number of nodes of the last \GC layer depends on the
dimension of the vector that we have to reconstruct.

The resulting 4-head network is represented in Figure~\ref{fig:final-network}, and it is trained by minimization of the
empirical risk given by $ w_\text{main}  \mathcal{R}_\text{main} + w_\text{AE}  \mathcal{R}_\text{AE} + w_\text{FR}
\mathcal{R}_\text{FR}^f + w_\text{ER}  \mathcal{R}_\text{ER}^f $. The parameters $\vartheta_\text{sh}$,
$\vartheta_\text{main}$, $\vartheta_\text{AE}$, $\vartheta_\text{FR}^\text{f}$, $\vartheta_\text{ER}^\text{f}$ are
trained by stochastic gradient descent. As in previous works \citep{kipf2016semi}, the gradient update is performed
batch-wise, using the full dataset for every training iteration. The proposed framework inherits memory and time
complexity from the underlying layers we chose to use for each of the functions $g_\text{sh}$, $g_\text{main}$,
$g_\text{AE}$, $g_\text{FR}^\text{f}$, $g_\text{ER}^\text{f}$. Thus, for the architecture here presented, it means a
memory complexity linear in the number of edges for a sparse representation of the adjacency matrix $\Matrix{A}$, and a
time complexity linear in the number of edges \citep{kipf2016semi}.

Note that by forcing some of $w_\text{AE}$, $w_\text{FR}^{f}$, and $w_\text{ER}^{f}$ to be identically equal to zero, we
can achieve with the same network architecture a settings where some of the auxiliary tasks are effectively deactivated,
and in the limit case where all of them are equal to zero we recover the standard \GCN.

%% file: results.tex
\section{Experimental Results}

\subsection{Datasets and Experimental Setup}
We test our models on semi-supervised classification using the standard datasets Citeseer, Cora, and Pubmed
\citep{sen2008collective}. These are citation networks, where graph vertexes correspond to documents and (undirected)
edges to citations. The vertex features are a bag-of-words representation of the documents. Each node is associated to a
class label. The Cora dataset contains 2.708 nodes, 5.429 edges, 7 classes and 1.433 features per node. The Citeseer
dataset contains 3.327 nodes, 4.732 edges, 6 classes and 3.703 features per node. The Pubmed dataset contains 19.717
nodes, 44.338 edges, 3 classes and 500 features per node.

For the training phase, we used the same setting adopted by \citet{kipf2016semi}, which in turns follows the
experimental setup of \citet{yang2016revisiting}. We allow for only 20 nodes per class to be used for training. However,
since this is a semi-supervised setting, the training procedure can use the features of all the nodes. The test and
validation sets comprise 1.000 and 500 nodes respectively. All the train/test/splitting used in the experiments are the
ones used by \citet{kipf2016semi}.

% \subsection{Experimental setup}
Since as already mentioned in Section~\ref{sec:final-network} we opted to use the \GC layers as foundational building
blocks, the fair baseline comparison is with {\GCN}s. We conducted three kinds of experiments. In the first round of
tests, we focused on one-hidden layer architectures with 16 units, so that we could compare these results directly with
the one presented in \citet{kipf2016semi}. For each of the three datasets, the networks that we compared against the
baseline and against each other are the ones showed in Table~\ref{tab:results-one-hidden}. Each of them are instances of the
multi-head architecture presented in Section~\ref{sec:final-network}, with some of the heads deactivated. We used the
same amount of dropout, L2 regularization, optimizer \citep{kingma2014adam}, and learning rate used by
\citet{kipf2016semi}, unless otherwise stated. The sub-tasks weights appearing in \Eqref{eq:multi_task_loss} have been
tuned for all the networks by means of grid search. For the network \emph{main} + \muacro{FR}, we tuned whether it is
better the full or the partial reconstruction and we found the optimal number of reconstructed features by searching
through the values 100, 200, 400, 800 for the Citeseer and Cora, and through the values 50, 100, 200 for Pubmed
\footnote{
    As an example, considering Citeseer, which has 3.703 dimensions, we searched through 100, 200, 400, 800
reconstructed features, meaning that we zero-ed out 3.603, 3.503, 3.303, 2.903 number of features, respectively.
}. 
The
resulting optimal values have been used also for the networks \emph{main} + \muacro{AE} + \muacro{FR} and  \emph{main} +
\muacro{AE} + \muacro{FR} + \muacro{ER}. Similarly, for the network \emph{main} + \muacro{FR} we tuned whether it is
better the full or the partial reconstruction and we found the optimal number of reconstructed embeddings by searching
through the values 2, 4, 8 for all the datasets. The resulting optimal values have been used also for the networks
\emph{main} + \muacro{AE} + \muacro{ER} and  \emph{main} + \muacro{AE} + \muacro{FR} + \muacro{ER} as well. Finally, we
tuned the sets $\Set{V}_\text{AE}$, $\Set{V}_\text{FR}$, $\Set{V}_\text{ER}$ by comparing the two limiting cases
$\Set{V}_\text{AE} = \Set{V}_\text{FR} = \Set{V}_\text{ER} = \Set{V}_\text{l}$ and $\Set{V}_\text{AE} =
\Set{V}_\text{FR} = \Set{V}_\text{ER} = \Set{V}$.

With Citeseer and Cora we trained for 5.000 epochs, while with Pubmed we trained for 2.500 epochs. During the training,
the learning rate was reduced by a factor 10 if the multi-task loss on the validation set did not improve for 40 epochs
in a row. In all the cases, we selected the best performing epoch on the validation set to assess the final performance
on the test set. Each network has been trained and tested 10 times, each time with a different random weights
initialization, and randomized reconstruction features and embeddings (if applicable).

In the second rounds of experiments we focused on the Cora dataset and we compared one, two and five hidden layers architectures. 
The goal was to assess if the
proposed framework allows to achieve good results even when increasing the network depth, thus making less relevant to
tune the number of hidden layers in {\GCN} architectures in order to reduce the ``over-smoothing'' phenomenon whenever
the networks become deeper. The experimental setup is the same as in the first round of experiments, and each hidden 
layer is a 16 units \GC layer.

Finally, in the third rounds of experiment, we compared our best results with the best results achieved by
\citet{you2020does}, who employed an approach based on self-supervision and multi-task learning.

\subsection{Results}
\begin{table}[!t]
    \caption{
        \label{tab:results-one-hidden}
        The table shows the mean classification accuracy achieved on the test sets and the standard error of the mean
        for the one-hidden layer networks on Citeseer, Cora, and Pubmed. The acronyms
        \muacro{AE}, \muacro{FR}, and \muacro{ER} refer to the three auxiliary tasks described in
        Section~\ref{sec:tasks}. For completeness, we show also the results reported by
        \citet{kipf2016semi}, which however do not report the standard errors. 
        }
    \vspace{-.5em}

    % \smallskip
    \centering
    \resizebox{.99\columnwidth}{!}{
    \begin{tabular}{lccccccc}

        \noalign{\smallskip}
        
        % \hline
        \cline{2-4}
        \noalign{\smallskip}
        
        & \multicolumn{3}{c} {\textbf{Accuracy}} \\
        
        \noalign{\smallskip}
        \hline
        \noalign{\smallskip}
        
        \textbf{Network} & \textbf{Citeseer} & \textbf{Cora} & \textbf{Pubmed} \\

        \noalign{\smallskip}
        \hline
        \noalign{\smallskip}
        
        \GCN \citep{kipf2016semi} 
            & $70.3 \%$ & $81.5 \%$ & $79.0 \%$ \\
        \GCN (our 10 runs) 
            & $69.84 \pm 0.22 \%$ & $81.13 \pm 0.13 \%$ & $78.63 \pm 0.41 \%$  \\

        \hdashline[1pt/3pt]
        \noalign{\smallskip}
        
        \emph{main} + \muacro{AE} 
            & $71.06 \pm 0.16 \%$ & $\bm{82.17\pm 0.09 \%}$ & $78.97 \pm 0.14 \%$ \\

        \emph{main} + \muacro{FR} 
            & $70.94 \pm 0.13 \%$ & $82.07 \pm 0.15 \%$ & $78.91 \pm 0.19 \%$ \\

        \emph{main} + \muacro{ER} 
            & $70.42 \pm 0.20 \%$ & $81.83 \pm 0.12 \%$ & $\bm{79.33 \pm 0.07 \%}$ \\

        \emph{main} + \muacro{AE} + \muacro{FR} 
            & $\bm{71.14 \pm 0.12 \%}$ & $82.13 \pm 0.10 \%$ & $78.92 \pm 0.14 \%$ \\

        \emph{main} + \muacro{AE} + \muacro{ER} 
            & $69.95 \pm 0.14 \%$ & $82.05 \pm 0.14 \%$ & $79.15 \pm 0.13 \%$ \\

        \emph{main} + \muacro{AE} + \muacro{FR} + \muacro{ER} 
            & $70.09 \pm 0.20 \%$ & $81.96 \pm 0.13 \%$ & $79.15 \pm 0.13 \%$ \\

        \noalign{\smallskip}    
        \hline
    \end{tabular}
    }
    % \vspace{-1.2em}
\end{table}

\begin{table}[!t]
    \caption{
        \label{tab:results-cora}
        The table shows the mean classification accuracy achieved on the test sets and the standard error of the mean
        for one/two/five-hidden layer networks on Cora. In parenthesis we show the increase in performance measured in
        percentage points (pp) with respect to baseline \GCN architectures. The acronyms \muacro{AE}, \muacro{FR}, and
        \muacro{ER} are the same as in Table~\ref{tab:results-one-hidden}.
        }
    % \smallskip
    \vspace{-.5em}
    \centering
    \resizebox{.99\columnwidth}{!}{
    \begin{tabular}{lccccccc}
        
        \noalign{\smallskip}
        
        \cline{2-4}
        \noalign{\smallskip}
        
        & \multicolumn{3}{c} {\textbf{Accuracy on Cora}} \\
        
        \noalign{\smallskip}
        \hline
        \noalign{\smallskip}
        
        \textbf{Network} & \textbf{1 hidden layer} & \textbf{2 hidden layers} & \textbf{5 hidden layers} \\

        \noalign{\smallskip}
        \hline
        \noalign{\smallskip}
        
        \GCN (our 10 runs) 
            & $81.13 \pm 0.13 \%$ & $79.74 \pm 0.54 \%$ & $16.47 \pm 2.19 \%$  \\

        \hdashline[1pt/3pt]
        \noalign{\smallskip}
        
        \emph{main} + \muacro{AE} 
            & $\bm{82.17\pm 0.09 \%}$ (+1.04pp) & $\bm{81.10 \pm 0.28 \%}$ (+1.36pp) & $49.57 \pm 5.50 \%$ (+33.10pp) \\

        \emph{main} + \muacro{FR} 
            & $82.07 \pm 0.15 \%$ (+0.94pp) & $80.89 \pm 0.29 \%$ (+1.15pp) & $\bm{51.05 \pm 4.32 \%}$ (+34.58pp) \\

        \emph{main} + \muacro{ER} 
            & $81.83 \pm 0.12 \%$ (+0.70pp) & $80.37 \pm 0.20 \%$ (+0.63pp) & $34.01 \pm 4.56 \%$ (+17.54pp) \\

        \emph{main} + \muacro{AE} + \muacro{FR} 
            & $82.13 \pm 0.10 \%$ (+1.00pp) & $80.85 \pm 0.25 \%$ (+1.11pp) & $49.51 \pm 5.58 \%$ (+33.04pp) \\

        \emph{main} + \muacro{AE} + \muacro{ER} 
            & $82.05 \pm 0.14 \%$ (+0.92pp) & $79.76 \pm 0.23 \%$ (+0.02pp) & $25.21 \pm 3.41 \%$ (+8.74pp) \\

        \emph{main} + \muacro{AE} + \muacro{FR} + \muacro{ER} 
            & $81.96 \pm 0.13 \%$ (+0.83pp) & $79.45 \pm 0.13 \%$ (-0.29pp) & $26.23 \pm 3.40 \%$ (+9.76pp) \\

        \noalign{\smallskip}    
        \hline
    \end{tabular}
    }
    % \vspace{-1.2em}
\end{table}

\begin{table}[!t]
    \caption{
        \label{tab:comparison}
        The table shows the increase in accuracy (measured in percentage points) compared to the \GC building block
        achieved by the best architectures of Table~\ref{tab:results-one-hidden} and those proposed by \citet{you2020does}. Note
        that our architectures produced a larger or comparable increase in accuracy.
        }
    \vspace{-.5em}

    % \smallskip
    \centering
    \resizebox{.9\columnwidth}{!}{
    \begin{tabular}{lccc}
        \cline{2-4}
        \noalign{\smallskip}
        & \multicolumn{3}{c} {\textbf{Increase in accuracy}} \\
        \noalign{\smallskip}
        \hline
        \noalign{\smallskip}
        \textbf{Network}            & \textbf{Citeseer} & \textbf{Cora} & \textbf{Pubmed} \\
        \noalign{\smallskip}
        \hline
        \noalign{\smallskip}
        Ours                        & \textbf{1.30 pp} & \textbf{1.04 pp} & 0.70 pp \\
        Best of \citep{you2020does} & 0.83 pp & 0.81 pp & \textbf{0.90 pp} \\
        \noalign{\smallskip}
        \hline
    \end{tabular}
    }
    % \vspace{-1.2em}
\end{table}

Table~\ref{tab:results-one-hidden} shows the results for the one-hidden layer architectures. It can be seen that
all the proposed self-supervised multi-task frameworks achieve better mean accuracy than a plain \GCN architecture.
Moreover, the best performing architecture in each dataset is always one of ours, with a corresponding improvement in
performance ranging from 0.70 to 1.30 percentage points. Interestingly, the best performing architecture always shows a
smaller standard error compared to the plain \GCN, thus exhibiting a more stable performance at different random weights
initialization.

The better performance verified with our one-hidden layer networks are confirmed in the second rounds of experiments
(see Table~\ref{tab:results-cora}). Also in this case, the best performing architectures are those proposed in this
paper, which additionally show a reduced variance compared to \GCN. These results suggest that the proposed
architectures help to alleviate the ``over-smoothing'' problem affecting deep {\GCN}s.

It is worth noticing that, the best performing architectures varies depending on the dataset at hand. In particular, the
full network with 4 active heads was never the best performing candidate model (but still better than the baseline).

Finally, in Table~\ref{tab:comparison} we compare our results with those obtained by the best architectures proposed in
\citep{you2020does}. Since the base building blocks in \citet{you2020does} achieve performance different than ours, and
considering that we used the same building block (\ie \GC layer), Table~\ref{tab:comparison} shows the increase (\ie
delta) in accuracy with respect to the building block, to keep the comparison fair. It is possible to notice that our
architectures produced a larger or comparable increase in accuracy.

%% file: conclusion.tex
\section{Conclusion}
We introduced three self-supervised auxiliary tasks to improve semi-supervised classification performance on graph 
structured data by training them in a multi-task framework. Precisely, 
\ie
\begin{enumerate*}[label=(\roman*)]
    \item vertex features autoencoding;
    \item corrupted vertex features reconstruction;
    \item corrupted vertex embeddings reconstruction.
\end{enumerate*}

The experiments we performed on standard datasets showed better performance with respect to {\GCN}s. Moreover, we
compared our results with those achieved by \citet{you2020does} and we showed a larger or comparable increase in
accuracy with respect to the base building block.

The two/five-hidden layers scenarios showed that the proposed architectures are more stable and can achieve better
results compared to the \GCN baselines. These considerations suggest that the proposed framework helps to alleviate the
``over-smoothing'' problem affecting deep {\GCN}s \citep{li2018deeper}.

A possible future work could be to replace the \GC layer with other neural layers devoted to deal with graph structured
data (\eg Graph Attention Networks \citep{gat}) to analyze advantages and drawbacks.